\documentclass[10pt,twocolumn,letterpaper]{article}

 \usepackage[pagenumbers]{cvpr}              

\definecolor{cvprblue}{rgb}{0.21,0.49,0.74}
\usepackage[pagebackref,breaklinks,colorlinks,allcolors=cvprblue]{hyperref}

\def\paperID{*****} 
\def\confName{CVPR}
\def\confYear{2025}

\title{TuneComp: Joint Fine-tuning and Compression for Large Foundation Models}

\author{Xiangyu Chen, Jing Liu, Ye Wang, Matthew Brand, Pu (Perry) Wang, Toshiaki Koike-Akino \\
{ Mitsubishi Electric Research Laboratories (MERL), Cambridge, MA 02139, USA}\\
{\tt\small \{xiachen, jiliu, yewang, brand, pwang, koike\}@merl.com}
}

\begin{document}
\maketitle
\begin{abstract}

To reduce model size during post-training, compression methods, including knowledge distillation, low-rank approximation, and pruning, are often applied after fine-tuning the model. However, sequential fine-tuning and compression sacrifices performance, while creating a larger than necessary model as an intermediate step.
In this work, we aim to reduce this gap, by directly constructing a smaller model while guided by the downstream task.
We propose to jointly fine-tune and compress the model by gradually distilling it to a pruned low-rank structure. Experiments demonstrate that joint fine-tuning and compression significantly outperforms other sequential compression methods.

\end{abstract}
\section{Introduction}

Large transformer-based models have demonstrated superior performance in many tasks, including natural language processing and computer vision. However, their enormous size requires substantial computational power and memory for fine-tuning and deployment to effectively adapt to downstream tasks. To tackle these, Parameter-Efficient Fine-Tuning (PEFT) was widely explored to reduce memory consumption in fine-tuning. In deployment, various post-training compression techniques have been employed in these fine-tuned models, including weight quantization, knowledge distillation, and network pruning. However, sequentially fine-tuning and compression often suffers from significant performance loss. 

\begin{figure}[ht]
\vskip 0.2in
\begin{center}
\centerline{\includegraphics[width=1\columnwidth]{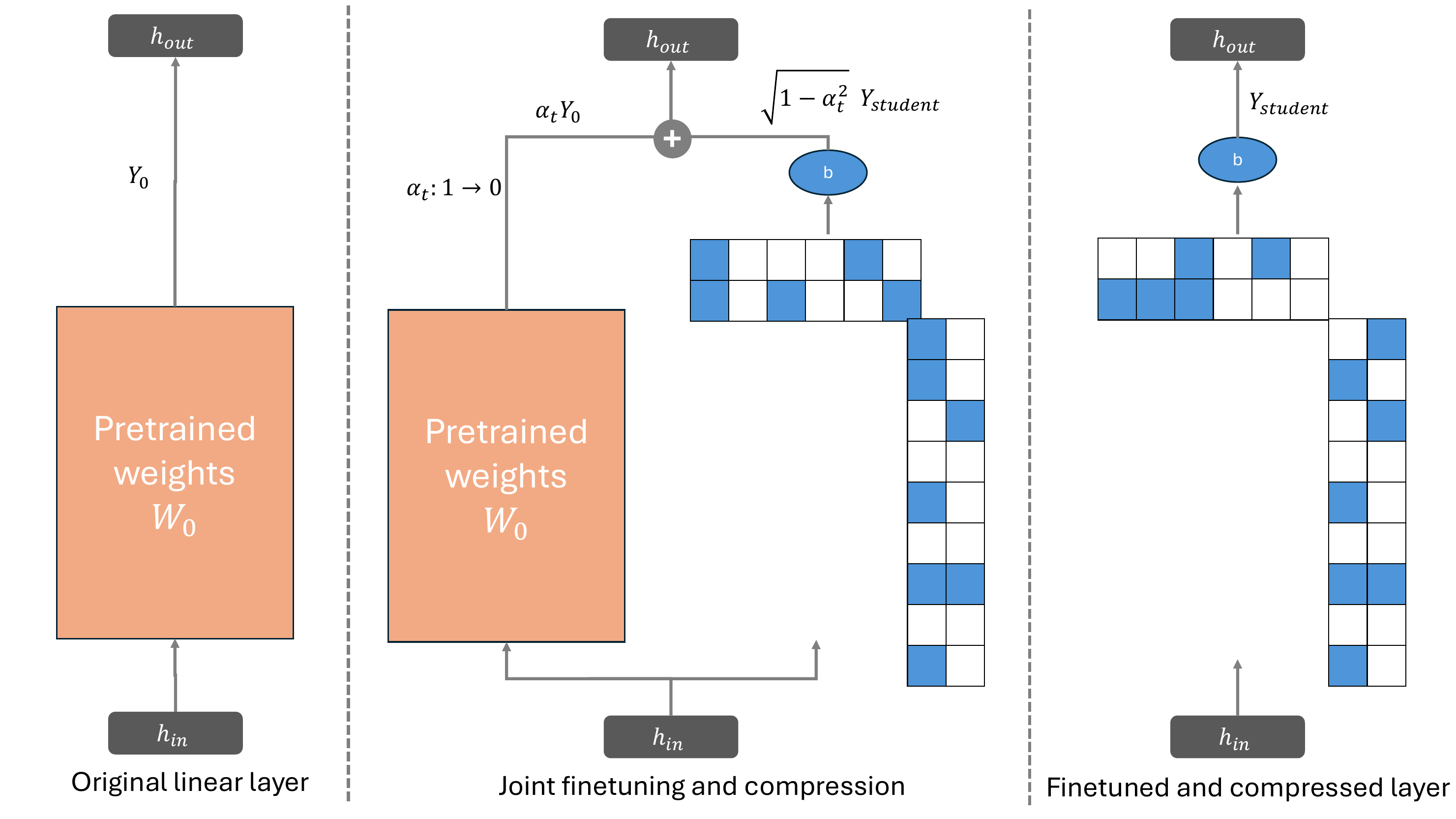}}
\caption{The proposed joint fine-tuning and compression pipeline, where compression involves low-rank approximation as well as the pruning/sparsification of the low-rank structures.}
\label{fig_illu}
\end{center}
\vskip -0.2in
\end{figure}
\begin{figure}[ht]

\begin{center}
\centerline{\includegraphics[width=0.9\columnwidth]{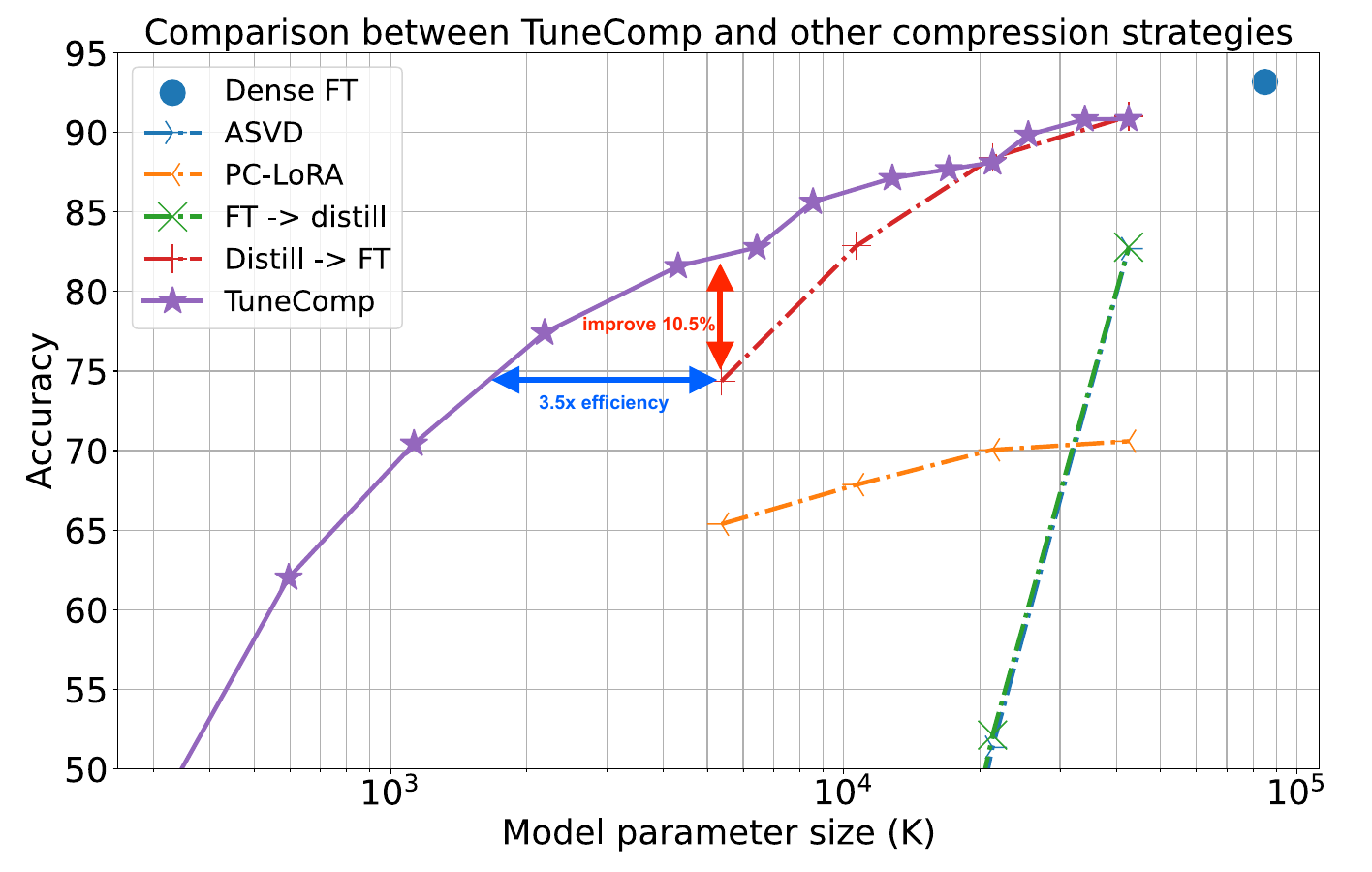}}
\caption{Comparison between our proposed TuneComp, jointly fine-tuning and compression, with other compression strategies.}
\label{fig_brand}
\end{center}
\vskip -0.35in
\end{figure}

Joint fine-tuning and distillation has emerged as a promising alternative~\cite{hwang2024pc}. This approach was first proposed in computer vision tasks to jointly fine-tune and prune convolutional neural networks iteratively~\cite{tung2017fine}.
Recently, PC-LoRA~\cite{hwang2024pc} uses a decay factor to reduce the influence of pretrained teacher branch smoothly, discarding the switching between fine-tuning and compression. The inclusion of fine-tuning in the compression process leads to much higher compression ratios, while maintaining comparable accuracy. However, the potential of the progressive compression pipeline is under-explored. The structure of the student branch with a basic low-rank decomposition can be further compressed with pruning techniques. Further, the performance can be enhanced through more careful design of the transition from the teacher to the student branch.

Extending the progressive compression pipeline, our work proposes \textbf{TuneComp} as shown in \cref{fig_illu}.
In this framework, each linear projection layer is split into two parallel branches. The first branch is the frozen weights of the pretrained model, while the second is the trainable and compressed student branch based on low-rank approximation.
During fine-tuning, the student branch gradually replaces the teacher branch, which reduces its influence on the output to zero, while maintaining the total power from both branches equal to 1.

We also employ novel extensions of
activation-aware initialization.
To further compress the model, pruning is jointly applied on top of this low-rank decomposition. 
The entire pipeline integrates fine-tuning, knowledge distillation, low-rank decomposition, and pruning, and directly producing a compact and efficient model. 

\vspace{-3pt}
\section{Related Work}
\subsection{Joint fine-tuning and compression}

From fine-tuning to compression, there are multiple pipelines to select \cite{lopes2024computer,cheng2024survey,chen2024comprehensive}. One widely used is fine-tuning downstream tasks first and then distilling to small models for deployment. In this path, researches either focus on improving fine-tuning to closing the gap between source domain of pretrained dataset and the new domain from downstream tasks, hence making the fine-tuned model generalize better on the given the tasks. For example, a notable line of work is Parameter-Efficient Fine-Tuning (PEFT), such as LoRA~\cite{hulora}, LoHA \citep{yeh2024navigating}, LoKr \citep{yeh2024navigating}, KronA \citep{edalati2022krona}, SuperLoRA \citep{chen2024superlora}, LoDA \citep{liu2023loda} and CorDA \cite{yang2024corda}. Another line aims to compress the post-training model to achieve similar performance to the original model while significantly reducing model size. Work in this line includes pruning \cite{bai2024sparsellm,isik2023gpt,frantar2023sparsegpt,men2024shortgpt,kwon2022fast}, parameter efficient structures \cite{yuan2023asvd,chenslaying,sehanobishstructured,wang2021pufferfish}, knowledge distillation \cite{kwonreward}, quantization \cite{lin2024awq,dettmers2022gpt3,yuan2023rptq,nagel2019data,banner2019post,cai2020zeroq} and combined \cite{ji2022neural,huang2022compressing,liberatori2022yolo}. Others compress the model first and fine-tune the small model for the target tasks \cite{hsu2022language,liu2019rethinking,su2020sanity,wang2020gan,yao2021joint,chen2021scatterbrain,xiong2021nystromformer,choromanski2021rethinking}. However, these sequential pipelines from whichever direction, fine-tune then compress or compress then fine-tune, introduced an intermediate agent, either the large fine-tuned model or the small compressed model, restricting the interplay between fine-tuning and compression, and often with significantly degraded performance. Therefore, some other works tried to combine finetuning and compression. However, this line of work is still underexplored. \cite{tung2017fine} proposes to fine-tune and compress the model alternatively, which is time-consuming and suboptimal. PC-LoRA \cite{hwang2024pc} eliminated the alternative switching between teacher and student model with a decay factor that gradually decreasing with time, thus gradually reducing the dependence on teacher model. 

\subsection{Low-Rank initialization}

Initialization of low-rank approximation, a parameter efficient structure $\mathbf{W} \approx \mathbf{B} \mathbf{A}$, has been studied in many scenarios. As adapters in parameter-efficient fine-tuning, LoRA \cite{hulora} proposed either $\mathbf{A}$ or $\mathbf{B}$ should be initialized as zero matrix, while the other one to be random Gaussian matrix.
While  \cite{hayou2024impact} argues that Gaussian initialization for the $\mathbf{A}$ matrix can lead to better performance by introducing more instability. Weight-aware initialization is also studied.
PiSSA \cite{meng2024pissa} instead initialized them with {\it{SVD}} decomposition. {{Nystrom initialization}} \cite{li2024crucial} combined both the pretrained matrix $\mathbf{W_0}$ with Gaussian initialization.
CorDA \cite{yang2024corda} further integrated covariance of activations, leading to activation-aware initialization.
However, all these methods target in the adapter-initialization in fine-tuning.
For compression, ASVD \cite{yuan2023asvd} also claimed that task information should be integrated in the initialization to emphasize different features given different tasks and hence proposed general activation-aware initialization based on the statistics of activations.
However, low-rank initialization in joint fine-tuning and compression lacks investigation, given PC-LoRA \cite{hwang2024pc} simply used Gaussian initialization for the low-rank matrices.
\section{Method}

\subsection{Basic low-rank approximation via SVD}
Low-rank approximation uses a pair of low-rank matrices $\mathbf{A}\in \mathbb{R}^{r\times d_\text{in}}$ and $\mathbf{B}\in \mathbb{R}^{d_\text{out}\times r}$ to approximate a matrix $\mathbf{W}\in \mathbb{R}^{d_\text{out}\times d_\text{in}}$, via
$    \mathbf{W} \approx \mathbf{BA}$, where typically the rank $r \ll \min\{d^\text{in}, d^\text{out}\}$. Compared to the original weight matrix $\mathbf{W}$, the parameters used for the low-rank approximation reduces from $d^2$ to $2dr$ when $d^\text{in}=d^\text{out} = d$. However, this approximation becomes less accurate with smaller rank.

A natural way to realize the low-rank approximation is to use the singular value decomposition (SVD), which decomposes a matrix $\mathbf{W}$ as
$    \mathbf{W} = \mathbf{USV}^T$,
where $\mathbf{U}$, $\mathbf{V}$ are the left and right singular vectors, respectively, and $\mathbf{S}$ is a diagonal matrix of non-negative singular values that are sorted in descending order, by convention.
Truncating the SVD to keep only the largest $r$ singular values and corresponding vectors, yields the optimal rank-$r$ approximation of $\mathbf{W}$ (in terms of minimum Frobenius norm error),
$    \mathbf{W} \approx \mathbf{U}_r\mathbf{S}_r\mathbf{V}_r^T$,
where $\mathbf{U}_r$ and $\mathbf{V}_r$ denote the first $r$ columns of $\mathbf U$ and $\mathbf V$, respectively, and $\mathbf{S}_r$ is the top-left $r \times r$ block of $\mathbf S$, corresponding to the largest (most important) singular values.
The low-rank approximation $\mathbf B \mathbf A$ can be formed by
\begin{align}
    \mathbf{B} = \mathbf{U}_r \mathbf{S}^{1/2}_r, \
    \mathbf{A} = \mathbf{S}^{1/2}_r \mathbf{V}^T_r.
\end{align}

\noindent \textbf{Notation:} we will use $\mathbf{S}_{-r}$ to denote the bottom-right $r \times r$ block of $\mathbf S$, and $\mathbf{U}_{-r}$ and $\mathbf{V}_{-r}$ to denote the corresponding last $r$ columns of $\mathbf U$ and $\mathbf V$, respectively.

\subsection{Activation-aware low-rank approximation}

Focusing on only approximating the pretrained weights is suboptimal as it does not consider the effect of the activation statistics on the overall performance of the model.
Ultimately, the objective is to approximate $\mathbf{WX}$, given a calibration set of $l$ sample activations arranged in the matrix $\mathbf{X} \in \mathbb{R}^{d_\text{in} \times l}$.
Following the methodology of~\cite{yuan2023asvd, wang2024svd}, we adopt an invertible transform matrix $\mathbf{C}$ to adapt the weights to the pattern of activations $\mathbf{X}$.
In this framework, SVD is performed on $\mathbf{WC}$, as given below:
\vspace{-2pt}
\begin{align}
\label{eq:asvd}
    \mathbf{W} &= \mathbf{(WC)C^{-1}} = \text{SVD}(\mathbf{WC})\mathbf{C}^{-1} = (\mathbf{USV}^{T})\mathbf{C}^{-1}
\end{align}

Keeping only the top $r$ singular values/vectors, we have the low-rank approximation
   $ \mathbf{W} \approx \hat{\mathbf{W}} := (\mathbf{U}_r\mathbf{S}_r\mathbf{V}_r^T)\mathbf{C}^{-1}$.
As derived by~\cite{wang2024svd}, an optimal choice, to minimize 
\begin{equation}\label{eq;activation_aware}
    \|\hat{\mathbf{W}}\mathbf{X} - \mathbf{W}\mathbf{X} \|_F^2
\end{equation}
is to set $\mathbf{C}$ equal to the Cholesky decomposition of the covariance $\mathbf{XX}^T$.
This choice of $\mathbf{C}$ serves to whiten the activations, i.e., $\mathbf{C}^{-1}\mathbf{X} (\mathbf{C}^{-1}\mathbf{X})^T = \mathbf I$.
Similarly, setting $\mathbf{C}$ to be the square root (in terms of the singular values) of the covariance, i.e., $\mathbf{C} = (\mathbf{XX}^T)^{1/2}$, also achieves the optimum, which we called \textbf{RootCorDA}, in contrast to CorDA~\cite{yang2024corda} that sets $\mathbf{C} = (\mathbf{XX}^T)$.

In RootCorDA, one can set $\mathbf B=\mathbf{U}_r$ and $\mathbf A=\mathbf{S}_r\mathbf{V}_r^T \mathbf C^{-1}$ with $\mathbf{C} = (\mathbf{XX}^T)^{1/2}$, which achieves the global optimal of \cref{eq;activation_aware} given $\hat{\mathbf{W}}=\mathbf{BA}$.

\subsection{Prune low-rank structure during distillation}

To further compress the low-rank structure, we propose to further simultaneously prune the low-rank matrices during the low-rank distillation process via hard shrinkage. Specifically, given the target pruning ratio $\rho$, in the forward process of each iteration, we dynamically calculate the corresponding pruning threshold for each low-rank matrix, and zero out elements with magnitudes smaller than the threshold.
For example, for the low-rank matrix $\mathbf{A}$, we have
$\text{prune}(\mathbf{A}) = \mbox{HardShrink}(\mathbf{A}, \rho)$, where the HardShrink function sets the smallest $\rho$ percent of the elements in $\mathbf{A}$ (in terms of the magnitude) to zero.

At the end of our joint fine-tuning and compression iterations, the original weight matrix $   \mathbf{W}$ is approximated/replaced by the product of two pruned low-rank matrices, i.e., $\text{prune}(\mathbf{B}) \text{prune}(\mathbf{A})$.

\subsection{Joint fine-tuning, distillation and compression}
\subsubsection{Distill from teacher branch to student branch}\label{branch_decay}
Similar to PC-LoRA~\cite{hwang2024pc}, a progressive compression pipeline is employed. Specifically, the output of each linear layer can be denoted as
$    \mathbf{Y} = \alpha\mathbf{Y}_\text{teacher} + \alpha'\mathbf{Y}_\text{student}$,
where $\alpha$ decreases from one to zero according to
\begin{equation}
\label{decrease}
\alpha_t = \begin{cases} 
1 - \sin \left(\frac{\pi t}{2 T}\right) & \text{if } t \leq T \\
0 & \text{otherwise}
\end{cases}
\end{equation}

where $t$ is the current iteration number, and $T$ is the number of decaying iterations, which is set as 80\% of total number of iterations. 
In PC-LoRA~\cite{hwang2024pc}, $\alpha' = 1$ and 
$    \mathbf{Y}_\text{teacher} = \mathbf{W}_0\mathbf{X} + \mathbf{b}_0,     \mathbf{Y}_\text{student} = \mathbf{WX} + \mathbf{b}$.

Instead, to maintain the total power from both branches to be equal to one, we propose to use
$
    \alpha_t' =  \sqrt{1-\alpha_t^2}$.

Our overall pipeline becomes
\begin{equation}
     \mathbf{Y} = \alpha_t \underbrace{( \mathbf{W}_0\mathbf{X} + \mathbf{b}_0)}_\text{teacher} +\sqrt{1-\alpha_t^2} \underbrace{(\mathbf{W}\mathbf{X} + \mathbf{b})}_\text{student},
\end{equation}
where decay factor $\alpha_t$ is set as in \cref{decrease}. 

\subsubsection{Layer-wise regularization}\label{regularization}
To generalize better on downstream tasks, a regularization term is added to the overall loss:
\begin{equation}
\label{loss}
     L_\text{total} = L_\text{task}\{y, \hat{y}\} + \gamma L_\text{feat}\{F_{t}, F_{s}\}
\end{equation}
where 
$    L_\text{feat}\{F_{t}, F_{s}\} = \frac{1}{m}\sum_{i = 1}^m \text{MSE}(F_{t_i}, F_{s_i})$.
Here $y$, $\hat{y}$ are the ground truth and outputs of the network, $F_{s_i}$ and $F_{t_i}$ are outputs of $i$-th linear layer from student and teacher model respectively.

Instead of setting $\gamma$ as a constant $0.2$ in~\cite{hwang2024pc}, we propose to decrease $\gamma$ from 1 to 0 during the iterations, similar to \cref{decrease}. Our goal is to provide more layer-wise guidance at the beginning of the distillation iterations, and gradually relax this guidance.

\section{Experiments}

To validate our joint fine-tuning and compression method, we evaluated it on the Vision Transformer (ViT)~\cite{dosovitskiy2020image} model for image classification. 

\subsection{Settings}

We evaluated our TuneComp pipeline on the transfer learning task from ImageNet1K~\cite{hayou2024impact} to CIFAR100~\cite{krizhevsky2009learning} using the ViT-Base model as in~\cite{chen2024superlora}. We use training set of CIFAR100 as the fine-tuning dataset, and test on the testing set of CIFAR100. Similar to~\cite{chen2024superlora}, OneCycleLR scheduler is used and learning rate ranges from $7 \times 10^{-5}$ to $3 \times 10^{-3}$, and the best classification accuracy is recorded. The rank of the low-rank matrices of the student branch is tested from $\{32, 64, 128, 256\}$. Note that smaller rank corresponds to higher compression rate.

\subsection{Comparison of compression strategies}
We compare the proposed TuneComp with some widely used pipelines:
\begin{itemize}
    \item \textit{Fine-tune}: baseline, no compression at all.
    \item \textit{Fine-tune $\xrightarrow{}$ distill}: first full (dense) fine-tune each linear layer, and then compress the model by distillation.
    \item \textit{Distill $\xrightarrow{}$ fine-tune}: distill to a small model first, and then fine-tune the small model.
    \item \textit{TuneComp} (joint fine-tune and compress): progressively distill, prune, and fine-tune simultaneously, where the pruning ratios are tested between 0 to 90\%.
    \item \textit{Distill only}: compression only without fine-tuning.
\end{itemize}

The accuracy of the \textit{Distill only} method is very poor (below 40\%) and hence omitted from the figure. 
As shown by the Pareto fronts plotted in \cref{fig_brand}, \textit{TuneComp} significantly outperforms baseline methods. {\textit{Distill$\xrightarrow[]{}$fine-tune}} performs second best. Further, while {\textit{FT$\xrightarrow[]{}$distill}} can even outperform the state-of-the-art joint fine-tuning and compression method PC-LoRA at low compression rate, it is still dominated by our proposed method.

\subsection{Effect of low-rank matrix initialization}
We evaluated different initialization methods for the low-rank matrices $\mathbf B$ and $\mathbf A$ listed below.

\noindent (a) For Gaussian initialization, we evaluated:

\begin{itemize}[label=\hspace{14pt}$\bullet$]
\item   $\mathbf{B = 0}$, $\mathbf{A} =$ Gaussian~\cite{hayou2024impact};
    \item $\mathbf{B} =$ Gaussian, $\mathbf{A} =$ Gaussian;
\end{itemize}
The results, found in \cref{t1: initialization}, show that for the joint fine-tuning and compression case, initializing both low-rank matrices as Gaussian results in better performance, especially when the rank is low.

\begin{table}
\captionsetup{skip=15pt} 
\caption{Comparison of the classification accuracies by different low-rank matrices initialization methods for TuneComp.}
\label{t1: initialization}
\vspace{-15pt}
\begin{center}
\begin{small}
\begin{sc}
\begin{tabular}{lccccr}
\toprule
initialization& $r=32$ & $r=64$ \\
\midrule
$\mathbf{B = 0}$, $\mathbf{A} =$ Gaussian    & 65.92& 69.07 \\
$\mathbf{B} =$ Gaussian, $\mathbf{A} =$ Gaussian    & 70.13& 73.14 \\ \hline \hline
$\mathbf{B = 0}$, $\mathbf{A} = \mathbf{W}_0 \times$ Gaussian & 72.73 & 74.18  \\ \hline\hline
$\mathbf{B} = \mathbf{U}_{-r}$, $\mathbf{A} = \mathbf{V}_{-r}$  &  70.91& 72.37 \\
$\mathbf{B} = \mathbf{U}_{-r}\mathbf{S}_{-r}$, $\mathbf{A} = \mathbf{V}_{-r}$ & 70.35& 68.40 \\
$\mathbf{B} = \mathbf{U}_{-r}$, $\mathbf{A} = \mathbf{S}_{-r}\mathbf{V}_{-r}$ & 73.01& 72.57 \\
$\mathbf{B} = \mathbf{U}_{-r}\mathbf{S}_{-r}^{1/2}$, $\mathbf{A} = \mathbf{S}_{-r}^{1/2}\mathbf{V}_{-r}$ & 70.93& 72.42 \\ \hline 
$\mathbf{B} = \mathbf{U}_{r}$, $\mathbf{A} = \mathbf{V}_{r}$   &  75.03& 79.61 \\
$\mathbf{B} = \mathbf{U}_r \mathbf{S}_r$, $\mathbf{A} = \mathbf{V}_r$ & 78.01& 83.73 \\
$\mathbf{B} = \mathbf{U}_r$, $\mathbf{A} = \mathbf{S}_r\mathbf{V}_r$ & 77.96& 83.82 \\
$\mathbf{B} = \mathbf{U}_r \mathbf{S}_r^{1/2}$, $\mathbf{A} = \mathbf{S}_r^{1/2} \mathbf{V}_r$ & 78.74& 83.89 \\ \hline\hline 
CorDA & 76.70 & 83.34  \\
RootCorDA (proposed) & $\textbf{78.99}$ & $\textbf{84.62}$  \\
\bottomrule
\end{tabular}
\end{sc}
\end{small}
\end{center}
\vskip -0.3in
\end{table}

\noindent (b) For weight-aware initialization, we compared with Nystrom initialization~\cite{li2024crucial} where $\mathbf{B = 0}$, $\mathbf{A} = \mathbf{W} \times$ Gaussian. We also compared with many variants of SVD based initializations listed in \cref{t1: initialization}. 

We can see from \cref{t1: initialization} that, for SVD based low-rank decomposition, selecting top $r$ principle vectors works better than least $r$ singular vectors, and is much better than simple Gaussian initialization and Nystrom initialization. Symmetrically assigning singular values into left and right singular vectors is often a good choice. 

\noindent (c) For the activation-aware initialization, we evaluated:
\begin{itemize}[label=\hspace{14pt}$\bullet$]
    \item CorDA initialization from~\cite{yang2024corda};
    \item Proposed RootCorDA initialization.
\end{itemize}
The last few rows in \cref{t1: initialization} shows that CorDA in~\cite{yang2024corda} cannot consistently outperform the weight-aware SVD initialization (Symmetric singular values with top-r singular vectors). However, our RootCorDA outperforms CorDA.

\subsection{The effect of pruning}
\begin{figure}
\label{fig2_prune_ours}
\vspace{-2pt}
\begin{center}
\centerline{\includegraphics[width=0.9\columnwidth,trim={160 15 1040 53}, clip]{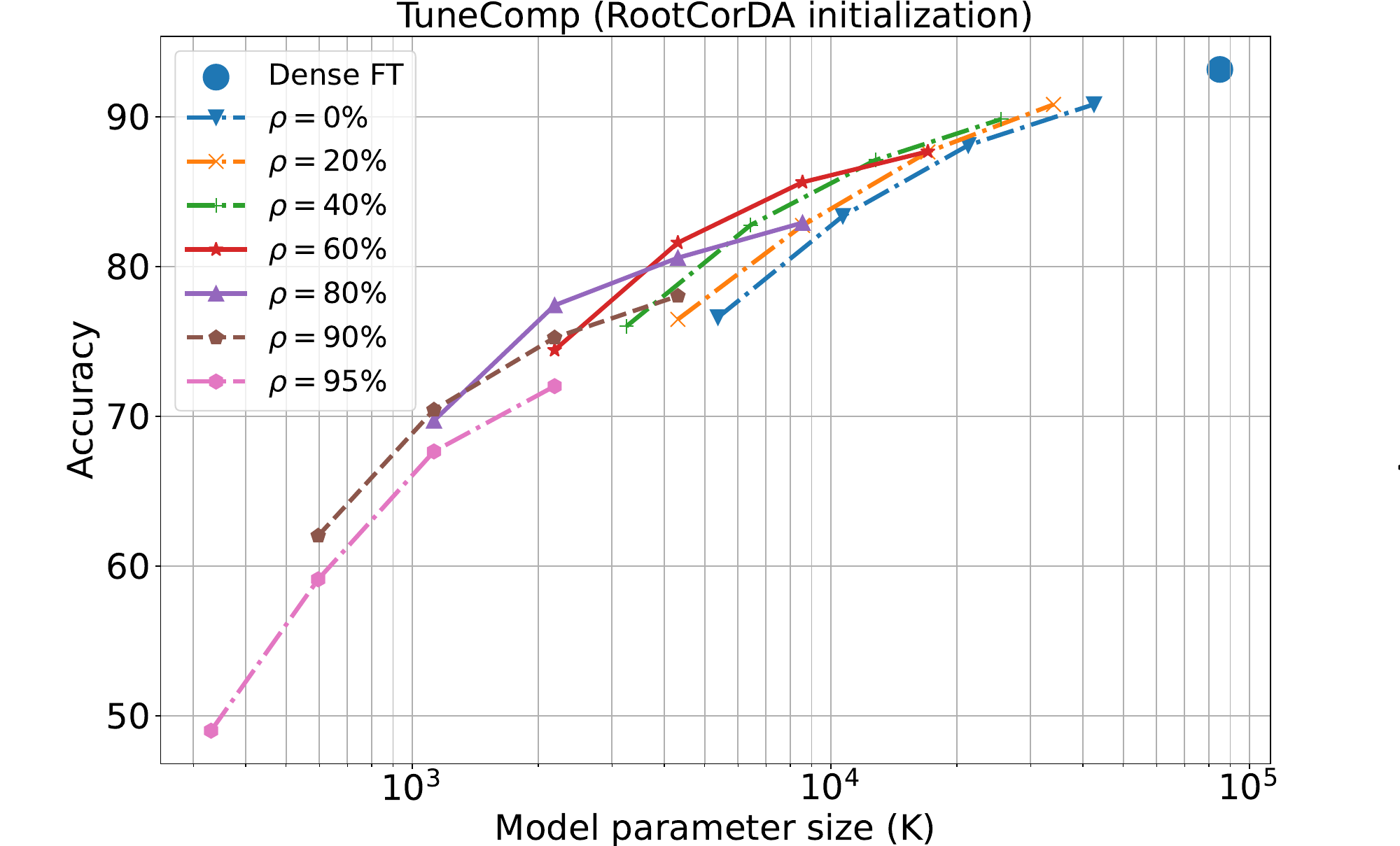}}
\caption{Performance of TuneComp under different pruning ratio $\rho \in \{0\%, 20\%, 40\%, 60\%, 80\%, 90\%, 95\%\}$.}
\label{fig2_prune}
\end{center}
\vskip -0.44in
\end{figure}

We studied the influence of integrating pruning into the compression. Specifically, during joint fine-tuning and distillation of the low-rank structure, pruning on the low-rank structure is simultaneously applied to each matrix $\mathbf{B}$ and $\mathbf{A}$, and the pruning ratio can be predefined. \cref{fig2_prune} demonstrates the performance of TuneComp under different pruning ratios. The blue dashed line corresponds to no pruning (ratio = 0\%), while other lines correspond to different pruning ratios. We can see that pruning 20\% (yellow dashed line) or 40\% (green dashed line) consistently outperforms the no pruning case, in terms of the accuracy-efficiency trade-off. However, pruning 60\% or more may not have a better accuracy-efficiency trade-off.

\subsection{Effect of regularization decay}
Finally, we compare our proposed regularization decay strategy in \cref{regularization} with the constant regularization $\gamma=0.2$~\cite{hwang2024pc}. The results are shown in \cref{t3: regularization}. It is clear that our proposed decay strategy achieves better performance.  

\begin{table}
\captionsetup{belowskip=10pt} 
\caption{Classification accuracy of TuneComp compressed model using different regularization weighting strategies. }
\label{t3: regularization}
\vspace{-12pt}
\begin{center}
\begin{small}
\begin{sc}
\begin{tabular}{lccccr}
\toprule
Regularization&  $r=32$ & $r=64$ & $r=128$ & $r=256$  \\
\midrule
constant ~\cite{hwang2024pc}& 78.39& 84.27& 88.12 & 90.80 \\ 
Dynamic & $\textbf{80.41}$ & $\textbf{86.11}$ & $\textbf{89.49}$ & $\textbf{91.69}$ \\
\bottomrule
\end{tabular}
\end{sc}
\end{small}
\end{center}
\vskip -0.3in
\end{table}
\section{Conclusion}
We proposed a method that simultaneously performs fine-tuning, knowledge distillation, low-rank approximation, and pruning. Experiments show that it significantly outperforms baselines, especially when compression rate is high. 

{
    \small
    \bibliographystyle{ieeenat_fullname}
    \bibliography{example_paper}
}


\end{document}